\documentclass[10pt,twocolumn,letterpaper]{article}

\usepackage{cvpr}              

\usepackage[dvipsnames]{xcolor}

\usepackage{multirow}
\definecolor{cvprblue}{rgb}{0.21,0.49,0.74}
\usepackage[pagebackref,breaklinks,colorlinks,citecolor=cvprblue]{hyperref}

\title{Invariance-powered Trustworthy Defense via Remove Then Restore}

\author{Xiaowei Fu, Yuhang Zhou, Lina Ma, and Lei Zhang\footnote{Corresponding Author} \\
Learning Intelligence \& Vision Essential (LiVE) Group\\
School of Microelectronics and Communication Engineering, Chongqing University, China\\
{\tt\small \{xwfu, yuhangzhou\}@cqu.edu.cn,} {\tt\small mln@stu.cqu.edu.cn,} {\tt\small leizhang@cqu.edu.cn}
}

\begin{document}
\maketitle
\begin{abstract}
Adversarial attacks pose a challenge to the deployment of deep neural networks (DNNs), while previous defense models overlook the generalization to various attacks. Inspired by targeted therapies for cancer, we view adversarial samples as local lesions of natural benign samples, because a key finding is that salient attack in an adversarial sample dominates the attacking process, while trivial attack unexpectedly provides trustworthy evidence for obtaining generalizable robustness. Based on this finding, a Pixel Surgery and Semantic Regeneration (PSSR) model following the targeted therapy mechanism is developed, which has three merits: 1) To remove the salient attack, a score-based Pixel Surgery module is proposed, which retains the trivial attack as a kind of invariance information. 2) To restore the discriminative content, a Semantic Regeneration module based on a conditional alignment extrapolator is proposed, which achieves pixel and semantic consistency. 3) To further harmonize robustness and accuracy, an intractable problem,
a self-augmentation regularizer with adversarial R-drop is designed. Experiments on numerous benchmarks show the superiority of PSSR.
\end{abstract} 
\section{Introduction}\label{intro1}
Adversarial attacks pose challenges to practical deployment of DNN based computer vision models. Adversarial defense \cite{goodfellow2014explaining,zhang2019theoretically,madry2017towards,zhou2021towards} are proposed to enhance the adversarial robustness against attacks. However, previous defense models often overfit known attacks and overlook the attack-invariance, an important factor for generalization. Recent studies on generalizability show that DNNs can learn invariance among various tasks \cite{yosinski2014transferable}. Therefore, a natural question is \textit{how to learn attack-invariant information}? To clarify this question, we have to answer another question: \textit{do attacks really share some common attributes}?


The targeted therapies for cancer inspire us. Tumors are often distributed in localized malignant lesions rather than the whole body before deterioration, while other areas always stay healthy. Targeted therapy can exactly kill the malignant tumor cells without affecting the surrounding normal cells. Similarly, some studies show that the perturbations tend to attack the fixed frequency band or have a specific attack preference \cite{huang2022adversarially,maiya2021frequency,karantzas2022understanding}. In other words, adversarial perturbations often attack the image locally rather than globally, which is similar to the locality prior of malignant tumor cells. Therefore, adversarial samples may be regarded as the ``canceration'' of benign samples to some extent, i.e., local lesions strike the healthy samples under artificial guidance, and the local lesions (salient poisoned pixels) often dominate the misclassification, while other areas (trivial poisoned pixels) are still relatively healthy distribution, but  unexpectedly show natural invariance. 
 This finding unveils the answer to above question: \textit{there really exist precious commonalities for attacks by decomposing into salient components and trivial components via pixel separation}.

\begin{figure}[t]
\centering
\includegraphics[width=0.95\columnwidth, height=0.6\columnwidth]{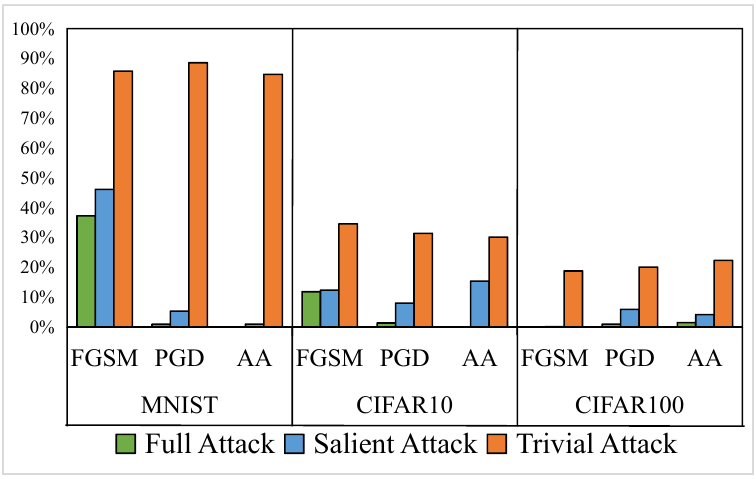} 
\caption{Attacking performances. The salient attacks dominate the wrong prediction comparable to the full attack, while the trivial attacks show significant adversarial robustness. 
}
\label{fig_verify}
\end{figure}

To confirm our findings, an exploratory experiment is conducted in Fig.~\ref{fig_verify}. Given an adversarial sample, we compute the contribution of each pixel to the final misclassification in terms of saliency map~\cite{baehrens2010explain,simonyan2013deep,erhan2009visualizing}, and set a threshold initialized as 1.5 times the mean value of the saliency map to decompose the salient and trivial perturbation from the total pixels. Intuitively, the pixels above the threshold are considered as \textit{salient} perturbations and the remaining are \textit{trivial} perturbations. Then, the salient components and trivial components are obtained by adding the trivial and salient perturbations to the benign sample. Popular FGSM, PGD and AA attackers are considered in this intriguing experiment on widely used MNIST and CIFAR10.

From Fig.~\ref{fig_verify}, we observe that 1) The salient attack dominates the full attack, contributes mostly to the misclassification, and represents the attack-specific information. 2) Trivial attacks do not seriously strike the benign semantics, and represent the invariance shared by various attacks, whose invariance is the self-consistency with the original healthy distribution. This validates our finding
about the local prior and pixel separation of adversarial perturbations. Based on the above findings, our first perspective is \textit{removing salient perturbation pixels can obtain adversarial but generalizable robustness}.

To explain the above perspective, we consider another fact that an image is composed of pixels, where each pixel is a kind of information carrier. Therefore, for an adversarial image, each pixel carries two kinds of information: attack noise and semantics, but not presented in a simple linear superposition due to isolation of pixels.
So it is impractical to divide each pixel into two subpixels. This is similar to the targeted therapy where the malignant tumor cells can only be either removed or retained. This supports our first perspective why to remove the saliently poisoned pixels. But a new question arises: \textit{since removing the saliently poisoned pixels improves robustness, whether the remaining trivial pixels carry sufficient discrimination (accuracy)}?

\begin{figure}[t]
\centering
\includegraphics[width=0.9\columnwidth, height=0.73\columnwidth]{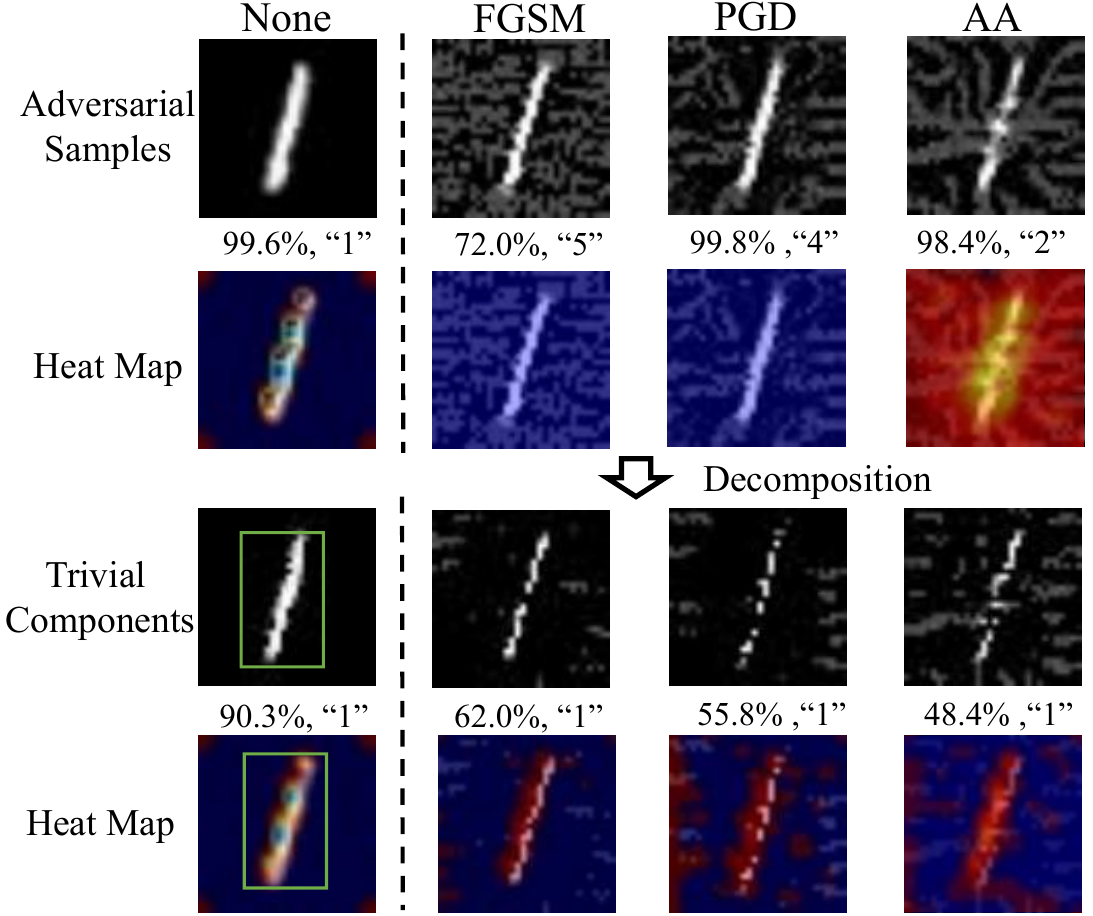} 
\caption{Trivial components and heatmaps are obtained by directly removing the salient pixels of each adversarial sample.}
\label{fig_heatmap}
\vspace{-0.4cm}
\end{figure}

We conduct another intriguing experiment, as shown in Fig.~\ref{fig_heatmap}, from which we observe: 1) the full attack strikes the original sample of digit ``1'' to false classes (``5'', ``4'', ``2''), but the trivial component always provides certain robustness with higher confidence on the true label ``1'' (i.e., 62.0\%, 55.8\% and 48.4\%) and correct semantics in attention around ``1'' (the 3rd row).
2) Although the robustness is improved, the discrimination of trivial components is degraded, since the saliently poisoned pixels also carry rich discriminative semantics.  This is observed by comparing the trivial components of adversarial samples to that of natural samples (confidence is degraded from 99.6\% to 62.0\%). Noteworthy that dropping pixels has almost no impact on the natural samples (confidence is degraded from 99.6\% to 90.3\% in the 1st column). This answers the above question that removing saliently poisoned pixels really improves robustness, but the semantic discrimination is degraded. Therefore, inspired by the healthy cell regeneration after targeted therapy for the cancer, our second perspective is \textit{regenerating the missing semantics from the trivial pixels is feasible to restore discrimination}.

Based on the key findings and perspectives, we propose to meet the challenge of adversarial but generalizable robustness from two-stage modules: Pixel Surgery (PS) and Semantic Regeneration (SR). PS aims to remove salient pixels for generalizable robustness. SR aims to restore discrimination from attack-invariant trivial pixels via a novel conditional alignment extrapolator.

\begin{figure*}[t]
\centering
\includegraphics[width=1.7\columnwidth, height=0.8\columnwidth]{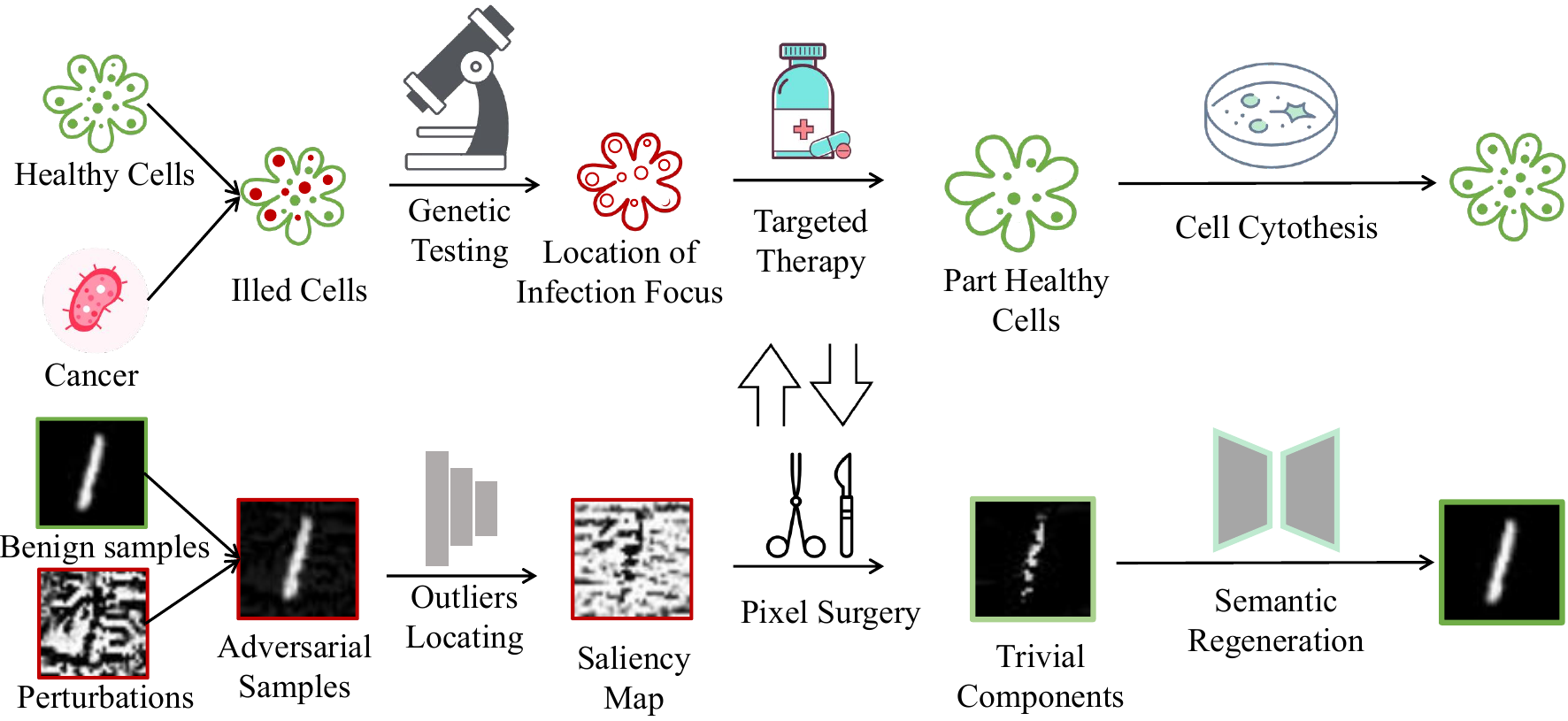} 
\caption{Adversarial sample is a local lesion of a benign sample and our PSSR is analogy to the targeted treatment and cell regeneration.}
\label{fig_cancer}
\end{figure*}

To sum up, inspired by the targeted therapy for cancers, we propose a Pixel Surgery and Semantic Regeneration (PSSR) model, as shown in Fig.~\ref{fig_cancer}.
The saliency maps conduct the ``genetic testing'' to adversarial samples to locate the malignant attacks, PS module equals to the ``targeted drug'' to provide generalizable robustness with attack-invariance, and SR module equals to a ``Petri dish'' to regenerate natural semantics. The contributions are two-fold:
\begin{itemize}
\item Inspired by targeted therapies, we have two key findings: 1) adversarial attacks can be decomposed into \textit{salient} and \textit{trivial} components at pixel level.
2) Salient components dominate adversarial attacks representing attack-specific information, while trivial components represent attack-invariance but lack discrimination.
\item Based on the observations and analysis, we propose a Pixel Surgery and Semantic Regeneration model. PS provides generalizable robustness by scoring the saliency maps and SR guarantees discrimination
via a Conditional Alignment Extrapolator.
To better reconcile robustness and accuracy, we further introduce a self-augmentation regularizer, Adversarial R-Drop.
\end{itemize}

\section{Related Work}

\subsection{Adversarial Attack}
Adversarial attacks mean the human imperceptible perturbations which could fool the model to make incorrect prediction. According to whether the attacker can access the target model, adversarial attacks can be generally divided into white-box attacks and black-box attacks.
White-box attacks mean attackers can fully access the target model, including model structure, parameters, gradients, etc. Several popular white-box attacks include gradient-based (e.g., FGSM \cite{goodfellow2014explaining}, BIM \cite{kurakin2016adversarial}, PGD \cite{madry2017towards}), classification layer perturbation based (e.g., DeepFool \cite{moosavi2016deepfool}), and conditional optimization based (e.g., CW \cite{carlini2017towards}, OnePixel \cite{su2019one}).
Black-box attacks mean an attacker has little prior knowledge about the target model and can be roughly divided into score-based, decision making-based, and transfer-based.
The score-based attack assumes the attacker can obtain one-hot prediction of the target model \cite{chen2017zoo}. Decision-based black-box attack assumes the attacker can only obtain the prediction of the model rather than scores (e.g., Boundary attacks \cite{brendel2017decision}). The transfer-based attack assumes the attacker cannot access any information of the target model \cite{dong2019evading,xie2019improving}, which are widely used for black-box attack.

\subsection{Adversarial Defense}
To improve adversarial robustness of DNNs, the defense models
can be roughly divided into input transformation \cite{xie2017mitigating}, model ensemble \cite{bagnall2017training}, metric learning \cite{zheng2016improving, bai2020adversarial}, certified defense \cite{raghunathan2018certified}, distillation defense \cite{Wang_2021_ICCV, papernot2016distillation}, and adversarial training \cite{goodfellow2014explaining,madry2017towards,kurakin2016adversarial,zhang2019theoretically}.

Image purification \cite{gu2014towards,vincent2008extracting,song2017pixeldefend} is another advanced defense strategy. The defender no longer needs to retrain the target model, but tries to purify the adversarial sample into a benign representation. This is more efficient since one cannot re-train the model for each new attack. \cite{meng2017magnet} trains a reformer network to move adversarial examples towards clean manifold. \cite{liao2018defense} trains a UNet that can denoise adversarial examples to their clean counterparts. \cite{samangouei2018defense} trains a GAN on clean examples and project the adversarial examples to the manifold of the generator. \cite{naseer2020self} trains a conditional GAN by playing a min-max game with a critic network. \cite{shi2021online} uses self-supervised learning to realize online image purification. 
Although the proposed PSSR restores natural semantics, it substantially differs from image purification based idea that we achieve salient attack degradation in pixel level and retain attack-invariant trivial pixels for semantic regeneration, from an explainable perspective of \textit{targeted therapies for cancers}. 

In addition, studies on explaining adversarial attacks are increasingly rich \cite{huang2022adversarially,maiya2021frequency,karantzas2022understanding,wang2020towards,naderi2022lpf} 
    from the perspective of frequency domain. However, these perspectives are not sufficiently explainable at the pixel level and cannot provide evidence of attack-invariance. Based on some intriguing experiments, we propose to decompose the adversarial perturbations into salient and trivial pixels.

\begin{figure*}[t]
\centering
\includegraphics[width=1.9\columnwidth, height=0.8\columnwidth]{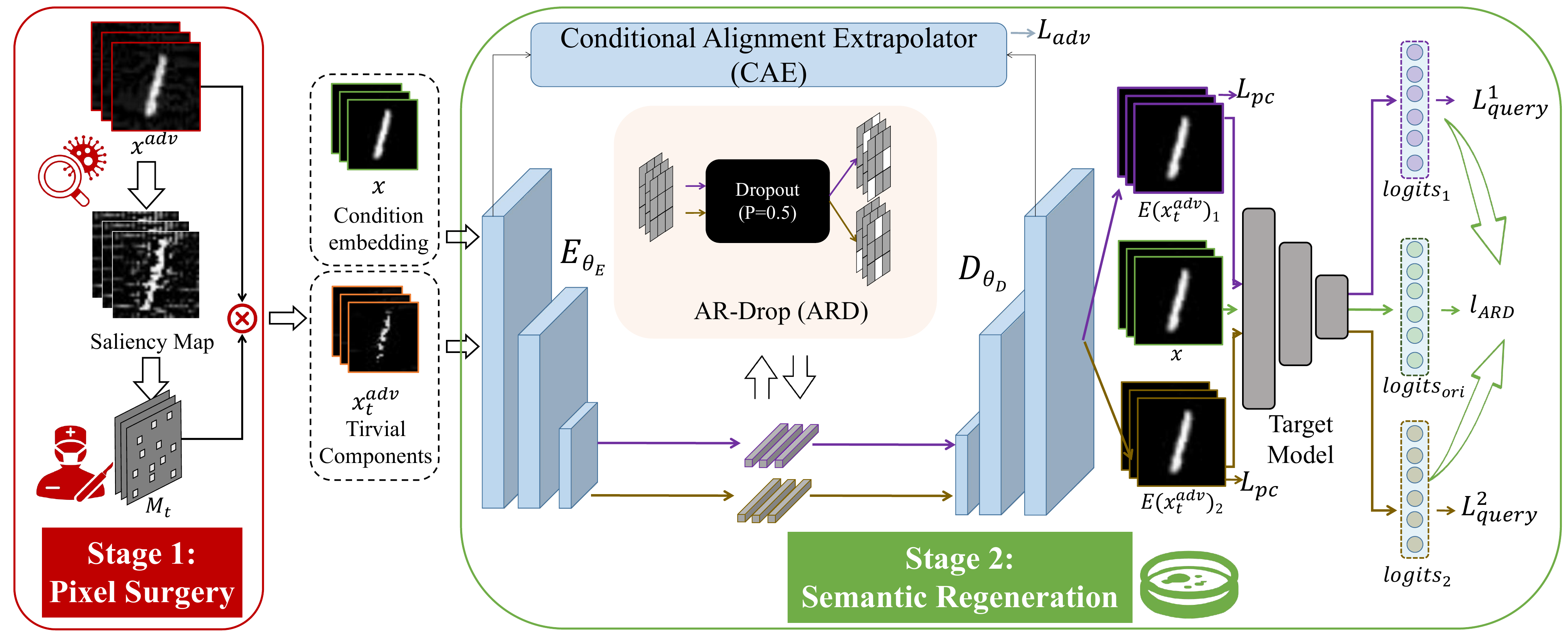} 
\caption{$\rm PSSR$ consists of two stages: Scoring-based pixel surgery (PS) and natural semantic regeneration (SR). SR consists of a conditional alignment extrapolator (CAE) and an adversarial R-Drop (ARD) regularizer. The input is not limited to adversarial samples.}
\label{fig_E2D}
\end{figure*}
\section{Methodology}
\subsection{Preliminary}
We focus on image classification task in this paper.
Let $x$  denote a natural image and $y$ denote its corresponding label, sampled from a joint distribution $D_{xy}$. 
By attacking a well-trained target model $F_w$ parameterized by $w$ 
an adversarial sample is generated by $x^{adv}=x+\delta^*$, in which $\delta^*$ represents the imperceptible perturbations for attack:
\begin{equation}
\begin{aligned}
\delta^*= \underset{\delta}{\arg \max } \ \   L\left(F_w(x+\delta, y)) \text {, s.t. } ||\delta||_{p} \leq \xi \right.
\end{aligned}
\label{eq1}
\end{equation}
where $L$ denotes a cross-entropy loss function,
the perturbation $\delta$ is bounded by $p$-norm within a $\xi$-ball (i.e., $||\delta||_{p} \leq \xi$), and $\xi$ represents the attack budget.

As the defender, we aim to maintain the model's robustness on the adversarial samples without losing accuracy on benign samples. This can be achieved either by adjusting the parameters $w$ with adversarial samples (i.e., adversarial training) or by purifying the adversarial sample $x^{adv}$ (i.e., adversarial purification):
\begin{equation}
\begin{aligned}
\underset{w/x^{adv}}{\min} L(F_w(x^{adv}, y))
\end{aligned}
\label{eq2}
\end{equation}
From the perspective of end-to-end training,
our model can be regarded as an image purification method, since the target model is kept frozen in PS and SR stages. 
\subsection{Pixel Surgery and Semantic Regeneration}

\textbf{Pixel Surgery.} A simple but effective pipeline of Pixel Surgery inspired by targeted therapies is presented in Fig.~\ref{fig_cancer} (Sec.~\ref{intro1}). We further  formalize it in this subsection.
We have analyzed that salient attacks dominate the attack process, which can be further explained by the finding in \cite{goodfellow2014explaining}, who argues that for approximately linear models, ``accidental steganography'' forces the network to pay much attention to the inputs having a greater inner product with model weights, thus resulting in erroneous but larger activation values. In other words, adversarial attacks tend to excessively maximize the $L_2$-norm of the network weights, resulting in abnormal activation values on some specific pixels (i.e., attack-specific salient pixels). Thus, we propose to directly remove such pixels by forcing them to zero.

Specifically, to locate such salient pixels, we can naturally compute the saliency map $S(x^{adv})$ \cite{baehrens2010explain,simonyan2013deep,erhan2009visualizing} of each adversarial sample as the following gradients:
\begin{equation}
\begin{aligned}
S(x^{adv})=\partial P(x^{adv}) / \partial x^{adv}
\end{aligned}
\label{eq3}
\end{equation}
where $P(x^{adv})$ is the prediction probability. The saliency map $S(x^{adv})$ describes the contribution of each pixel to final prediction through the change of different inputs on the final outputs. Then, the salient mask $\mathbf{M}_s$ and trivial mask $\mathbf{M}_t$ of each sample can be easily obtained via thresholding:
\begin{equation}
\begin{aligned}
M_s(x^{adv})_{i, j}=\left\{\begin{array}{l}
1, \quad S(x^{adv})_{i, j}\geq \epsilon \\
0, \quad \text {otherwise}
\end{array}\right.
\end{aligned}
\label{eq4}
\end{equation}
\begin{equation}
\begin{aligned}
\mathbf{M}_t(x^{adv})= \textbf{1} - \mathbf{M}_s(x^{adv})
\end{aligned}
\label{eq5}
\end{equation}
where $\textbf{1}$ is a matrix whose elements are all 1 having the same size as $\mathbf{M}_s(x^{adv})$ and $\epsilon$ is a threshold set defined as
\begin{equation}
\begin{aligned}
\epsilon=\alpha\cdot\overline{S}(x^{adv})
\end{aligned}
\label{eq_theshold}
\end{equation}
where $\overline{S}(x^{adv})$ is the mean value of $S(x^{adv})$ and $\alpha$ is a hyper-parameter, discussed in experimental section. 

Pixel surgery on a sample (not limited to adversarial samples) is then enforced to obtain the salient and trivial components by:
\begin{equation}
\begin{aligned}
x_s^{adv} = x^{adv} \cdot \mathbf{M}_s,~
x_t^{adv} = x^{adv} \cdot \mathbf{M}_t
\end{aligned}
\label{eq6}
\end{equation}

Preamble observations and analysis by revisiting Fig.~\ref{fig_heatmap} show that with the salient pixels removed, the retained trivial components provide generalizable robustness, but the discrimination is lost along with the removal of salient attacks. Therefore, a ``Petri dish" for trivial components is required to regenerate the discriminative natural semantics.

\begin{figure}[t]
\centering
\includegraphics[width=1\columnwidth, height=0.25\columnwidth]{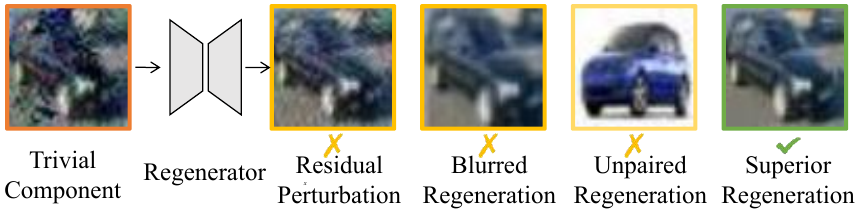} 
\caption{CAE outputs ``\textbf{True}'' if and only if the regenerated samples meet three necessary conditions, i.e., without residual perturbation, blurred and unpaired extrapolations.}
\label{fig_CGAN}
\end{figure}

\textbf{Semantic Regeneration. }To regenerate the original healthy distribution from the trivial components, we propose a Conditional Aligned Extrapolator (CAE) as the Petri dish.  Ideally, each regenerated sample shall meet three necessary conditions: 1) \textit{there is no residual adversarial perturbation}, 
2) \textit{it is paired with the corresponding benign sample}, and 3) \textit{it is consistent with the semantic cognition of the target model}. Technically, we take the conditional generating adversarial network (CGAN) \cite{mirza2014conditional} as the backbone of the proposed CAE regenerator, including an Extrapolator $E_{\theta_{E}}$ and a Discriminator $D_{\theta_{D}}$. For training the CAE regenerator, the trivial components are fed as input and the corresponding original sample serves as the conditional embedding of the extrapolator. If and only if the extrapolated data meet the above three conditions, the discriminator outputs \textbf{True}, otherwise \textbf{False}, as is shown in Fig.~\ref{fig_CGAN}. The proposed CAE follows an adversarial learning paradigm. The objectives are expressed as follows:
\begin{equation}
\begin{aligned}
L_{a d v}&=\underset{\theta_{E}}{\operatorname{min}} \ \underset{\theta_{D}}{\operatorname{max}} \ \mathbb{E}_{ x^{adv}_t, x}\left[\log D\left(x^{adv}_t, x\right)\right]\\ & +  \mathbb{E}_{x, x_t^{adv}}\left[\log \left(1-D\left(x^{adv}_t, E\left(x, x_t^{adv}\right)\right)\right)\right]
\end{aligned}
\label{eq7}
\end{equation}
where $x_t^{adv}$ represent the trivial pixels computed in Eq.~\ref{eq6}, $E$ and $D$ represents exptrapolator and discriminator, resp.

Furthermore, to ensure the pixel-level consistency and the semantic accuracy from the extrapolator $E$, we introduce a $L_1$-norm constrained pixel consistency loss  $L_{pc}$ and a cross-entropy based query loss $L_{query}$ to CAE as:
\begin{equation}
\begin{aligned}
L_{pc}(E)=\min \mathbb{E}\left[\| x-E\left(x, x_t^{adv}\right)\|_1\right]
\end{aligned}
\label{eq8}
\end{equation}
\begin{equation}
\begin{aligned}
L_{query}\left(E, F_w\right)=\min -\mathbb{E}\left[y \cdot \log \left(F_w\left(E\left(x_t^{adv}, x\right), w\right)\right)\right]
\end{aligned}
\label{eq9}
\end{equation}

Although CAE can meet the necessary conditions, it has to face another common but intractable challenge, i.e., trade-off between robustness (w.r.t. adv. samples) and accuracy (w.r.t. benign samples)~\cite{zhang2019theoretically}. If not reconciled, a leading problem is \emph{although SR enables the prediction confidence of correct label w.r.t. the adv. sample larger than the false label, it may still make the prediction of adv. sample inconsistent with the benign sample}. This is mainly due to that label only-based supervision tends to learn a strong mapping relationship, but overlooks the consistency between regenerated and benign samples.


To reconcile the robustness and accuracy, inspired by dropout~\cite{srivastava2014dropout} , 
we introduce a simple but effective Adversarial R-drop regularizer (ARD) in CAE. 
To prepare, stochastic self-augmentation sample pairs can be naturally obtained by randomly dropping out the feature of the inputs twice based on the Dropout operator. To improve the CAE regenerator, we strengthen the consistency between the outputs of the self-augmentation sample pairs and their corresponding benign sample, resp., via the ARD regularization:
\begin{equation}
\begin{aligned}
L_{ARD} & = KL(F_w(E(x^{adv}_t)_1), F_w(x)) \\& +KL(F_w(E(x^{adv}_t)_2), F_w(x))
\end{aligned}
\label{eq11}
\end{equation}
where $E(x^{adv}_t)_1$ and $E(x^{adv}_t)_2$ are naturally obtained by feeding $E(x^{adv}_t)$ from extrapolator into the Dropout module twice. Thus, the overall loss of CAE can be formalized as:
\begin{equation}
\begin{aligned}
L_{CAE} = L_{adv} + \lambda_{1}\cdot L_{query} + \lambda_{2}\cdot L_{pc} + \lambda_{3}\cdot L_{ARD}
\end{aligned}
\label{eq10}
\end{equation}
where $\lambda_{1}$, $\lambda_{2}$ and $\lambda_{3}$ are regularization coefficients, set to 1, 1 and 0.5, resp. by default.

\begin{figure}[t]
\centering
\includegraphics[width=0.8\columnwidth, height=0.9\columnwidth]{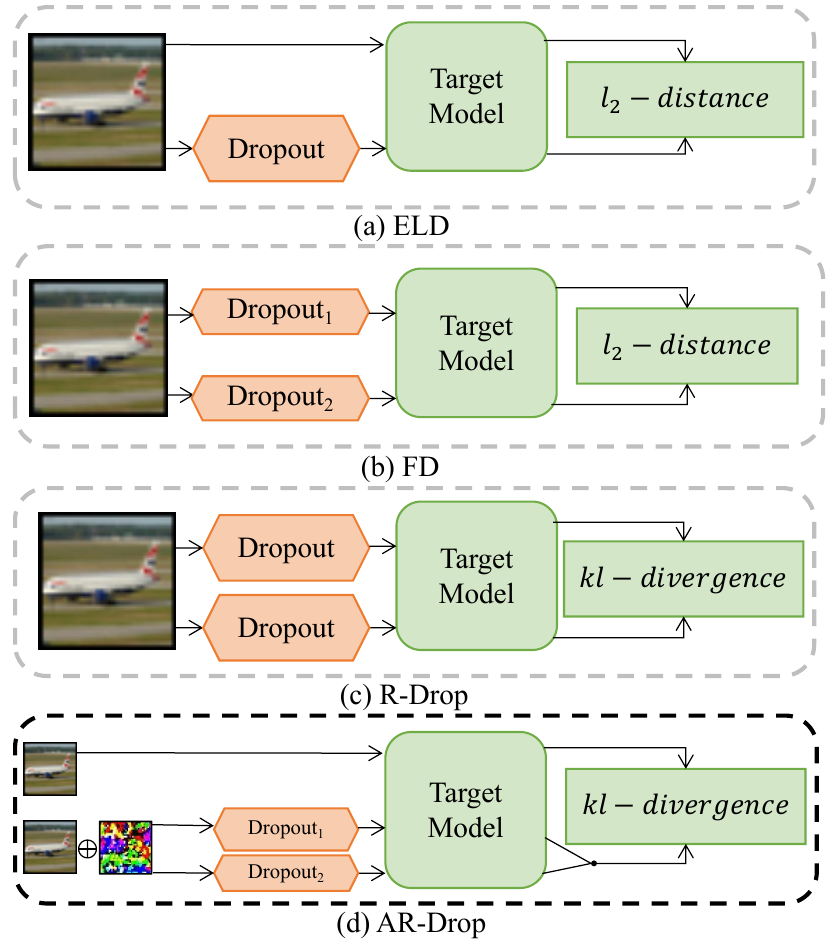} 
\caption{Structures of different Re-Dropout tricks.} 
\label{fig_ARD}
\end{figure}

\textbf{Discuss}: Re-Dropout based tricks are not proposed for the first time, e.g., ELD \cite{ma2016dropout}, FD \cite{zolna2017fraternal} and R-Drop \cite{wu2021r}. Yet, ELD directly reduces the gap between the sub-model with dropout (train) and the expected full model without dropout (inference). Both ELD and FD use the $L_2$ distance regularization on hidden states. R-Drop does not consider the trade-off between robustness and accuracy. In comparison, our AR-Drop both utilizes the distribution of multiple sub-models (with dropout) and reconciles accuracy and robustness. Their intuitive differences are shown in Fig.~\ref{fig_ARD}.

\textbf{Final Model. }Integrating the above three modules to obtain the final PSSR (Fig.~\ref{fig_E2D}), which consist of two stages:
\begin{itemize}
\item In PS stage, the salient pixels are removed to weaken the attack for certain robustness, and the retained trivial pixels unexpected provide attack-invariance.

\item In SR stage, trivial components are fed into the proposed CAE for healthy semantics regeneration, trained in an adversarial learning manner by Eq.~\ref{eq10}.
\end{itemize}

\begin{table*}[t]
\setlength{\tabcolsep}{3mm}
\renewcommand\arraystretch{0.9}
\begin{center}
\begin{tabular}{|c|c|ccccccc|}
\hline
 & Defense            & Benign & FGSM  & BIM   & MIM   & PGD   & AA  & CW   \\ \hline \hline
\multirow{7}{*}{\begin{tabular}[c]{@{}l@{}}MNIST\end{tabular}}
 & Baseline                                  & 99.30  & 39.72 & 0.56 & 0.58   & 0.48  & 0.00 & 0.56 \\
 & Grad Reg \cite{ross2018improving}                                   & 99.22  & 48.51 & 31.13 & 11.83 & 22.58 & 7.57 & 86.98  \\
 & $\rm TRADES_{(1\backslash\lambda=6.0)}$\cite{zhang2019theoretically}    & 98.79  & 95.47 & 94.67 & 94.33 & 96.07 & 95.22  & 99.16 \\
 & $\rm AT_{PGD}$ \cite{madry2017towards}                             & 98.70  & 92.15 & 95.99 & 98.71 & 96.41 & 95.83  & 98.09 \\
 & $\rm ARN_{PP}$ \cite{zhou2021towards}                            & 98.84  & 98.91 & 98.87 & \textbf{99.01} & 98.15 & \textbf{98.72} & 98.55  \\
 & $\rm PCL_p$  \cite{mustafa2020deeply}                             & 99.30  & 96.50 & 92.10 & 93.00  & 93.90  & -      & 91.30 \\
 & \textbf{PSSR(ours)}                  & \textbf{99.30}  & \textbf{99.15} & \textbf{98.90} & 98.85 & \textbf{98.92} & 98.69  & \textbf{99.22} \\ \hline \hline
\multirow{8}{*}{\begin{tabular}[c]{@{}l@{}}CIFAR10\end{tabular}}
 & Baseline                                   & 88.89 & 12.62 & 4.49 & 6.63 & 1.37 & 0.00 & 7.41 \\
 & Grad Reg\cite{ross2018improving}                                   & 82.20  & 20.75 & 4.37  & 6.43  & 1.54  & 1.23  &  35.79\\
 & $\rm TRADES_{(1\backslash\lambda=6.0)}$\cite{zhang2019theoretically}    &58.89  & 56.61 & 56.26 & 52.84 & 81.24 & 74.61 & 80.23 \\
 & $\rm AT_{PGD}$ \cite{madry2017towards}                            & 63.24  & 65.72 & 60.77 & 57.45 & 57.70 & 54.88 & 63.18 \\
 & $\rm ARN_{PP}$ \cite{zhou2021towards}                            & \textbf{91.79}  & 65.12 & 63.16 & 64.41 & 61.34 & 61.06 & \textbf{88.53} \\
 & Semi-SL \cite{uesato2019labels}                                    & 62.30  & 67.71 & 64.48 & 63.79 & 64.44 & 67.79 & 63.83 \\
 & $\rm PCL_p$ \cite{mustafa2020deeply}                               & 90.45  & 67.70 & 64.68 & - & 58.50 & -     & 73.70 \\
 & \textbf{PSSR(ours)}                  & 81.56   & \textbf{84.87} & \textbf{71.30} & \textbf{79.35} & \textbf{86.03} & \textbf{82.48} & 80.99 \\ \hline
\multirow{5}{*}{\begin{tabular}[c]{@{}l@{}}CIFAR100\end{tabular}}
 & Baseline   & \textbf{60.34}  & 1.53 & 1.21  & 16.78 & 2.71  & 0.00  & 12.54 \\
 & RO\cite{rice2020overfitting}         & 37.50  & - & 22.09  & -  & 23.83  & 25.45  & 22.19 \\
 & TRADES\cite{zhang2019theoretically}     & 52.59  & 39.04 & 31.62  & 30.71  & 28.55  & 31.26  & 31.40 \\
 & BagT\cite{pang2020bag}       & 38.89  & - & 31.44  & -  & 31.21  & 31.40  & 33.16 \\
 & \textbf{PSSR(ours)}       & 44.54  & \textbf{42.98} & \textbf{32.01}  & \textbf{44.48}  & \textbf{40.12}  & \textbf{31.74}  & \textbf{44.38} \\
 \hline
\end{tabular}
\end{center}
\vspace{-5.0mm}
\caption{Comparisons of $\rm PSSR$ with previous defense models under white-box attacks. 
The best result for each attack is shown in bold.}
\label{tab_white}
\vspace{-1.0mm}
\end{table*}

\begin{table*}[t]
\setlength{\tabcolsep}{3mm}
\renewcommand\arraystretch{0.9}
\begin{center}
\begin{tabular}{|c|c|ccccccc|}
\hline
 & Defense            & Benign & FGSM  & BIM   & MIM   & PGD   & AA   & CW  \\ \hline \hline
\multirow{6}{*}{\begin{tabular}[c]{@{}l@{}}MNIST\end{tabular}}
 & Baseline                                   & 99.30 & 45.06 & 0.59 & 0.76 & 0.60 & 0.26  & 30.07 \\
 & Grad Reg \cite{ross2018improving}                                  & 99.22  & 45.91 & 10.70  & 11.41  & 20.76 & 0.33  & 85.57 \\
 & $\rm TRADES_{(1\backslash\lambda=6.0)}$\cite{zhang2019theoretically}    & 98.79  & 95.35 & 94.54 & 94.19 & 97.15 & 94.90 & 90.06 \\
 & $\rm AT_{PGD}$\cite{madry2017towards}                             & 98.70  & 90.99 & 93.94 & 96.47 & 97.23 & 96.63 & 86.63 \\
 & $\rm PCL_p$\cite{mustafa2020deeply}                                & 99.30  & 78.30 & 72.70 & 74.50 & 69.50 & -     & 81.90 \\
 & \textbf{PSSR(ours)}                  & \textbf{99.30}  & \textbf{98.79} & \textbf{98.55} & \textbf{98.60} & \textbf{98.95} & \textbf{98.50} & \textbf{90.45} \\ \hline \hline
\multirow{6}{*}{\begin{tabular}[c]{@{}l@{}}CIFAR10\end{tabular}}
 & Baseline                                   & 88.89 & 11.43 & 4.02 & 6.73 & 1.52 & 3.27 & 0.00 \\
 & Grad Reg\cite{ross2018improving}                                   & 82.20  & 21.30 & 14.37  & 16.43  & 11.54  & 3.49  & 14.62 \\
 & $\rm TRADES_{(1\backslash\lambda=6.0)}$\cite{zhang2019theoretically}    & 57.21  & 56.87 & 58.90 & 54.21 & 77.75 & 70.19 & 76.51 \\
 & $\rm AT_{PGD}$ \cite{madry2017towards}                            & 63.24  & 53.78 & 57.33 & 53.91 & 53.88 & 51.22 & 56.68 \\
 & $\rm PCL_p$\cite{mustafa2020deeply}                                & \textbf{90.45}  & 85.50 & \textbf{83.70} & 81.90 & 76.40 & -     & 83.30 \\
 & \textbf{PSSR(ours)}                  & 81.56  & \textbf{86.15} & 83.38 & \textbf{83.66} & \textbf{78.31} & \textbf{72.76} & \textbf{83.91} \\ \hline

\multirow{4}{*}{\begin{tabular}[c]{@{}l@{}}CIFAR100\end{tabular}}
 & Baseline   & \textbf{60.34}  & 1.25 & 8.57  & 15.37 & 1.49  & 10.64  & 20.50 \\
 & $\rm AT_{PGD}$\cite{madry2017towards}         & 37.32  & 18.20 & 15.98  & 16.42  & 24.03  & 16.68  & 37.30 \\
 & TRADES\cite{zhang2019theoretically}         & 52.59  & 29.06 & 24.64  & 24.73  & 25.69  & 21.29  & 21.55 \\
 & \textbf{PSSR(ours)}         & 44.54  & \textbf{31.54} & \textbf{28.54}  & \textbf{30.30}  & \textbf{28.89}  & \textbf{21.86}  & \textbf{38.89 }\\
 \hline
\end{tabular}
\end{center}
\vspace{-5.0mm}
\caption{Comparisons of $\rm PSSR$ with previous defense models under black-box attacks. 
The best result for each attack is shown in bold.}
\label{tab_black}
\vspace{-5.0mm}
\end{table*}

\section{Experiments}


\subsection{Datasets and Backbones} We verify the effectiveness of $\rm PSSR$ on three widely used datasets, i.e, MNIST, CIFAR10 and CIFAR100.

$\bullet$ \textbf{MNIST} has 10 classes of handwritten digit images containing 60,000 training images
and 10,000 test images, and LeNet5 is adapted as the backbone.

$\bullet$ \textbf{CIFAR-10} has 10 classes of common objects containing 50,000 training images and 10,000 test images. ResNet18 \cite{he2016deep} is adapted as the backbone.

$\bullet$ \textbf{CIFAR-100} has 100 classes. Each class has 600 color images, including 500 training images and 100 test images. DenseNet121 \cite{huang2017densely} is adapted as the backbone.

\begin{table}[t]
\centering
\begin{tabular}{|c| c c c|}
\hline
 Defense & MNIST & CIFAR10  & CIAFR100\\
\hline
PSSR & 94.67 & 52.98 & 23.83 \\
\hline
\end{tabular}
\vspace{-2.0mm}
\caption{The performance of PSSR on adaptive BPDA attack.}
\label{bpda}
\vspace{-5.0mm}
\end{table}

\begin{table*}[]
\begin{center}
\setlength{\tabcolsep}{3.5mm}{
\renewcommand\arraystretch{0.9}
\begin{tabular}{|c|c|ccccccc|}
\hline
                   & Modules              & Benign & FGSM  & BIM   & MIM   & PGD   & AA  & CW    \\ \hline \hline
\multirow{5}{*}{MNIST} & Baseline         & 99.30  & 39.72 & 0.56  & 0.58  & 0.48  & 0.00  & 0.56 \\
                   & +PS                  & 99.30  & 62.22 & 40.61 & 38.49 & 31.24 & 15.36 & 34.67 \\
                   & +CAE                 & 98.27  & 98.87 & 95.71 & 96.79 & 96.94 & 95.07 & 97.94 \\
                   & +PS\&CAE             & 98.06  & 98.95 & 96.40 & \textbf{98.88} & 97.56 & 96.89 & 97.62 \\
                   & +PS\&CAE\&ARD        & \textbf{99.30}  & \textbf{99.15} & \textbf{98.90} & 98.85 & \textbf{98.92} & \textbf{98.69} & \textbf{99.22} \\ \hline \hline
\multirow{5}{*}{CIFAR10} & Baseline              & \textbf{88.89}  & 12.62 & 4.49  & 6.63  & 1.37  & 0.00  & 0.00 \\
                   & +PS                  & 88.89  & 27.81 & 26.84 & 25.73 & 26.41 & 21.12 & 29.96 \\
                   & +CAE                 & 77.32  & 69.52 & 62.78 & 68.55 & 77.10 & 73.05 & 76.67 \\
                   & +PS\&CAE             & 78.94  & 81.00 & \textbf{71.46} & 78.54 & 85.62 & 71.11 & 80.07\\
                   & +PS\&CAE\&ARD        & 81.56  & \textbf{84.87} & 71.30 & \textbf{79.35} & \textbf{86.03} & \textbf{82.48} & \textbf{83.91} \\ \hline
\multirow{5}{*}{CIFAR100} & Baseline      & \textbf{60.34}  & 1.53 & 1.21  & 16.78  & 2.71 & 0.00  & 12.54 \\
                   & +PS                  & 58.09  & 22.51 & 21.09 & 21.41 & 20.86 & 21.04 & 30.07\\
                   & +CAE                 & 41.98  & 33.88 & 25.66 & 24.74 & 36.68 & 28.67 & 43.80\\
                   & +PS\&CAE             & 42.13  & 41.15 & 30.73 & 41.28 & 38.58 & 30.45 & 41.76 \\
                   & +PS\&CAE\&ARD        & 44.54  &\textbf{42.98} & \textbf{32.01}  & \textbf{44.48}  & \textbf{40.12}  & \textbf{31.74}  & \textbf{44.38} \\
\hline
\end{tabular}}
\end{center}
\vspace{-5.0mm}
\caption{Ablation Analysis of the proposed PSSR, including PS, CAE and ARD. 
Each component contributes significant increment. }
\label{tab_ablation}
\vspace{-5.0mm}
\end{table*}

\begin{figure}[!t]
\centering
\includegraphics[width=1\columnwidth, height=0.4\columnwidth]{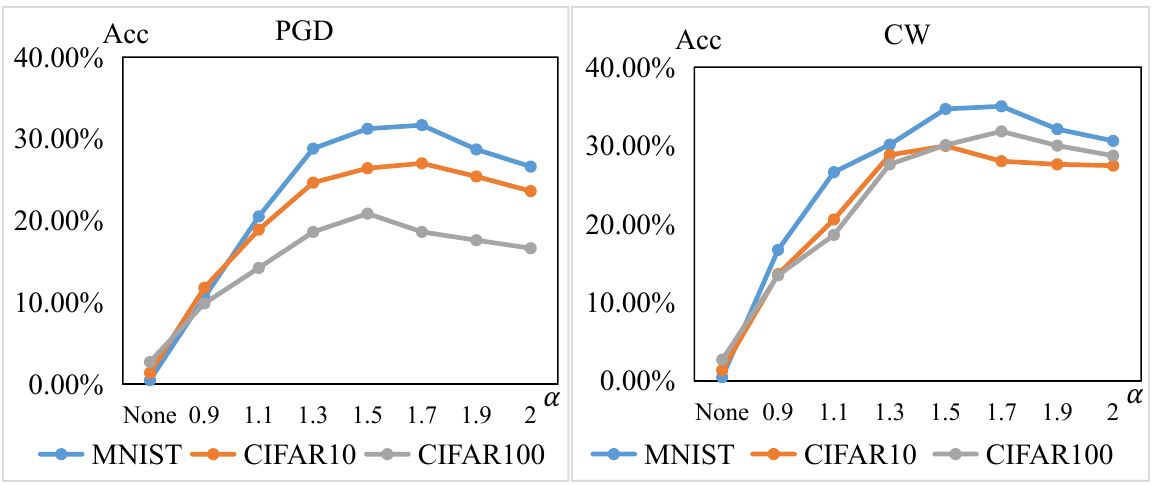} 
\caption{Selection of $\epsilon$ in Eq.\ref{eq_theshold}, in which $\alpha$ is discussed to reflect the impact of threshold, because $\overline{S}(x^{adv})$ can be pre-computed.}
\label{fig_hyper}
\vspace{-4.0mm}
\end{figure}

\subsection{Adversarial Attack Methods}
We consider the following attackers in our experiments.

\textbf{Fast Gradient Sign Method (FGSM)} \cite{goodfellow2014explaining} is a classic one-step adversarial attack based on gradient.
\textbf{Basic Iterative Attack (BIM)} \cite{kurakin2016adversarial} can be considered as an iterative variant of FGSM.
\textbf{Momentum Iterative Attack (MIM)} \cite{dong2018boosting} brings momentum acceleration to BIM.
\textbf{Projected Gradient Descent (PGD)} \cite{madry2017towards} is the standard evaluation for adversarial robustness, which brings projection and random initialization to BIM.
\textbf{Auto Attack (AA)} \cite{croce2020reliable} is the state-of-the-art attack, consisting
of an ensemble of parameter-free adversarial attacks such as auto-PGD \cite{croce2020reliable}, FAB \cite{croce2020minimally}, and Square Attack \cite{andriushchenko2020square}.
\textbf{CW Attack (CW)} \cite{carlini2017towards} is an optimization-based attack.

\subsection{Implementation Details} The target model is trained by the Adam optimizer for 200 epochs, and the initial learning rate is 0.001. The proposed PSSR is trained with 100 epochs 
and the initial learning rate is also 0.001. We use the implementation codes of FGSM, BIM, MIM and PGD in the advertorch toolbox \cite{ding2019advertorch}, and Auto-Attack \cite{croce2020reliable} released by the developers. FGSM and PGD can be regarded as the lower and upper bound of gradient-based attacks respectively, and used for joint adversarial training to better mine the attack-invariant information \cite{zhou2021towards}. Perturbation budgets are set as  $\xi=0.3$ and $\xi=8/255$ for MNIST and CIFAR10/CIFAR100, resp.

\subsection{Main Results to Advanced Attacks}
We test the robustness of PSSR to various attacks including known attacks (FGSM, BIM, MIM and PGD) and unknown attacks (AA and CW) under white-box and black-box attacking settings resp., and compare it with the mainstream and advanced defense methods in Table~\ref{tab_white} and Table~\ref{tab_black}. The results for other defense models are copied from their original papers under the same protocol for fairness. Note that for the black-box attack, following the traditional setting, we train a substitute model having the same loss function as the target model, which is considered to be a very effective black-box attack \cite{zhang2019theoretically}. Further, to explore the influence of obfuscated gradients, we deploy the proposed PSSR against BPDA~\cite{obfuscated-gradients} as an adaptive attack. The results in Table~\ref{bpda} show that the PSSR method could retain defense capacity on BPDA.
From the results, we observe that the proposed PSSR achieves comparable robustness (i.e., classification accuracy) for different attacks. This verifies the effectiveness of our idea.

\subsection{Ablation Analysis}

\textbf{Analysis of Different Components}. To delve into the effect of each component, such as Pixel Surgery (PS), Conditional Alignment Extrapolator (CAE) and the AR-Drop (ARD), we conduct ablation study 
under the white-box attack, as shown in Table~\ref{tab_ablation}. As expected, PS initially weakens the powerful attacks and provides a relatively friendly distribution (trivial pixels) for semantic regeneration via CAE. CAE extrapolates benign features from the trivial components to further restore semantics. ARD further improves the trade-off between accuracy and robustness.

\textbf{Analysis of the PS Threshold in Eq.~\ref{eq_theshold}}. The impact of threshold $\epsilon$ is shown in Fig.~\ref{fig_hyper}. According to Eq.~\ref{eq_theshold}, the threshold depends on a hyper-parameter $\alpha$ and the mean value $\overline{S}(x^{adv})$. Because $\overline{S}(x^{adv})$ can be pre-computed as a constant, we show the impact of $\alpha$ (0.9$\sim $2) to reflect $\epsilon$. We observe the trivial component achieves the best robustness for each dataset when $\alpha=1.5$, no matter for the gradient-based attack (PGD) or the optimization-based attack (CW). This shows the activations of natural samples are relatively uniform, and it is reasonable to choose a slightly larger threshold than the mean value of $S(x^{adv})$, by setting $\alpha$=1.5 to sufficiently remove powerful attacks (i.e., salient pixels).

\subsection{Interpretability of PSSR}
Improving adversarial robustness and interpretability of DNNs are equally important.
We discuss the interpretability of PSSR in improving robustness and give an in-depth analysis on the proposed three conditions in CAE. 

\begin{figure}[t]
\centering
\includegraphics[width=0.9\columnwidth, height=0.6\columnwidth]{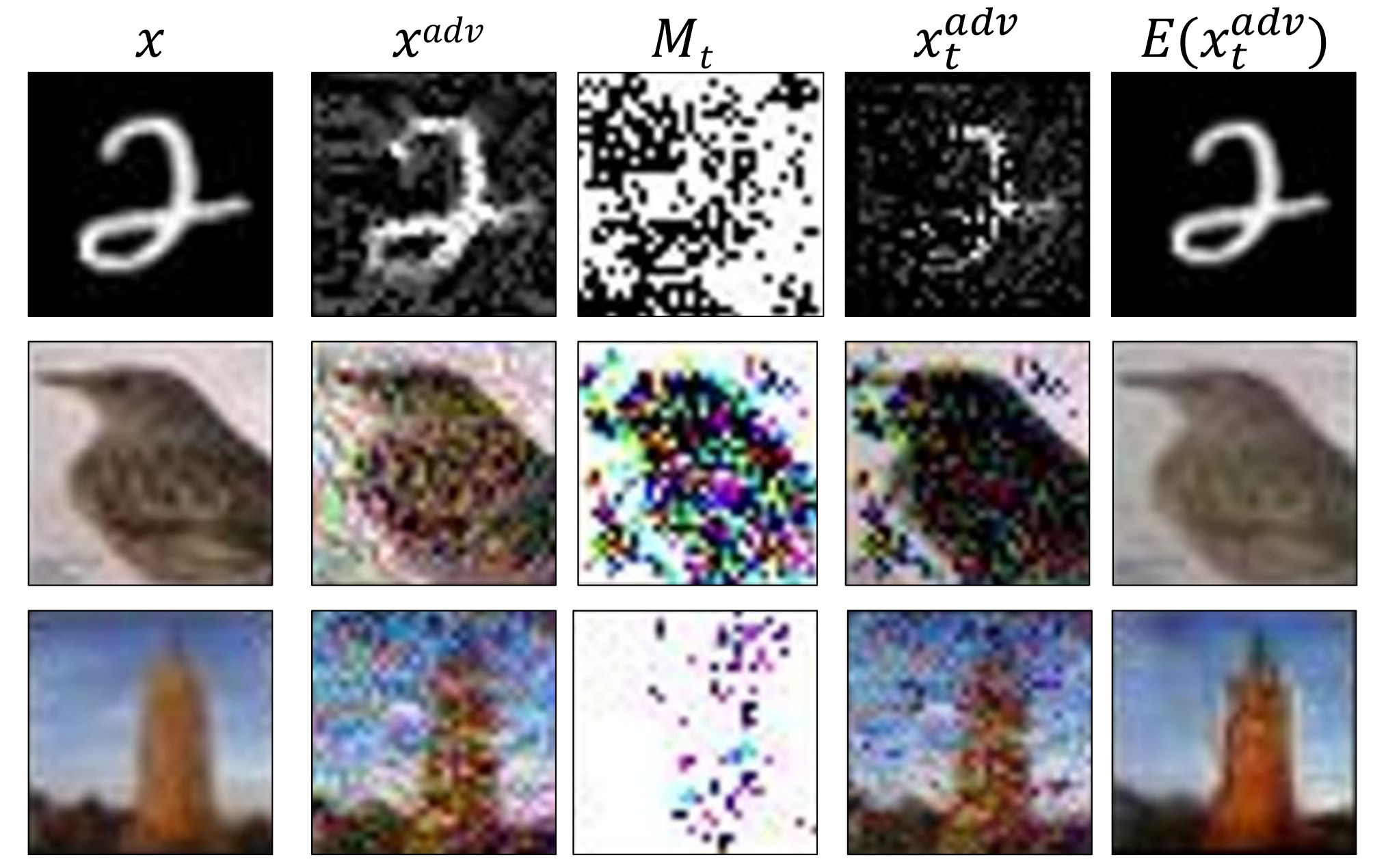} 
\caption{Visualization of the trivial components $x_t^{adv}$ and the regenerated samples $E(x_t^{adv})$ for three images sampled from MNIST, CIFAR10 and CIFAR100 resp. PGD attack is used.}
\label{fig_extra}
\vspace{-0.6cm}
\end{figure}

\textbf{Extrapolation Visualization.} To show the surgery and regeneration process of PSSR, Fig.~\ref{fig_extra} visualizes the benign$\backslash$adversarial sample ($x\backslash x^{adv}$), the trivial mask $\mathbf{M}_t$, the trivial components $x_t^{adv}$ and the regenerated sample $E(x^{adv}_t)$.
From Fig.~\ref{fig_extra}, we observe that
1) samples with large image capacity contain more salient outliers. For example, the trivial masks of CIFAR10 and CIFAR100 are more complex than that of MNIST.
2) The masks of CIFAR10 and CIFAR100 are distributed around the original semantics. This is explainable since adversarial attacks tend to change the semantic part of the original image, so the salient pixels of the perturbations are more concentrated on the original semantic. That is, the salient pixels also carry rich semantics being unavoidably removed together with adversarial outliers. Since $\mathbf{M}_t$ aims to retain trivial components, it is naturally distributed in peripheral area of the original semantics. 3) The proposed CAE meets our expectations on the first and second necessary conditions: \textit{paired with the natural sample $x$} and \textit{attack-free}.  The third condition is discussed in following feature visualization.

\begin{figure}[t]
\centering
\includegraphics[width=1.0\columnwidth, height=0.6\columnwidth]{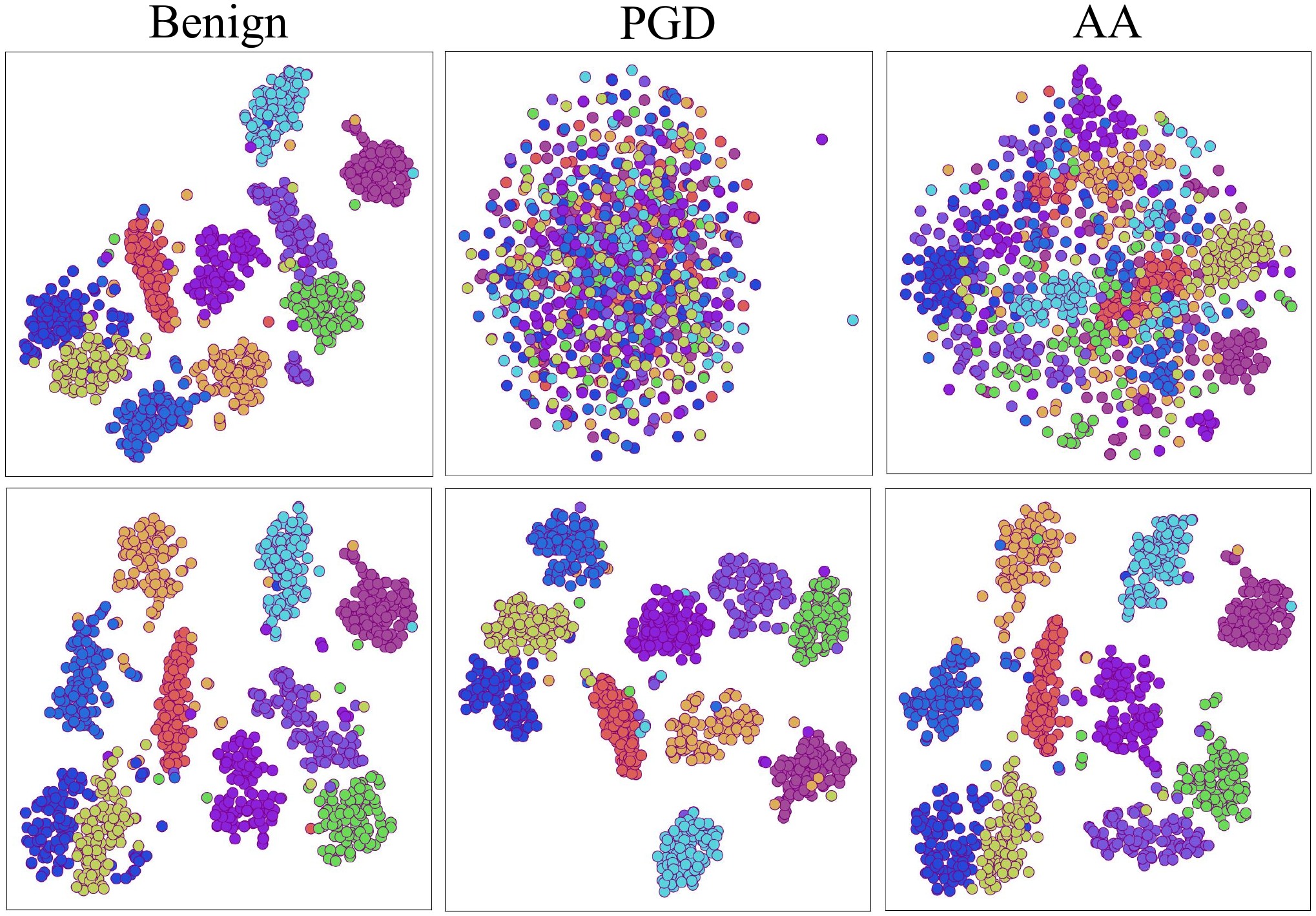} 
\caption{t-SNE visualization of feature distribution on CIFAR10: target model (the $\rm 1^{st}$ row) and PSSR model (the $\rm 2^{nd}$ row). 
The advanced AA and PGD attack are used for visualization.}
\label{fig_tsne}
\vspace{-4.0mm}
\end{figure}

\begin{figure}[t]
\centering
\includegraphics[width=1\columnwidth, height=0.4\columnwidth]{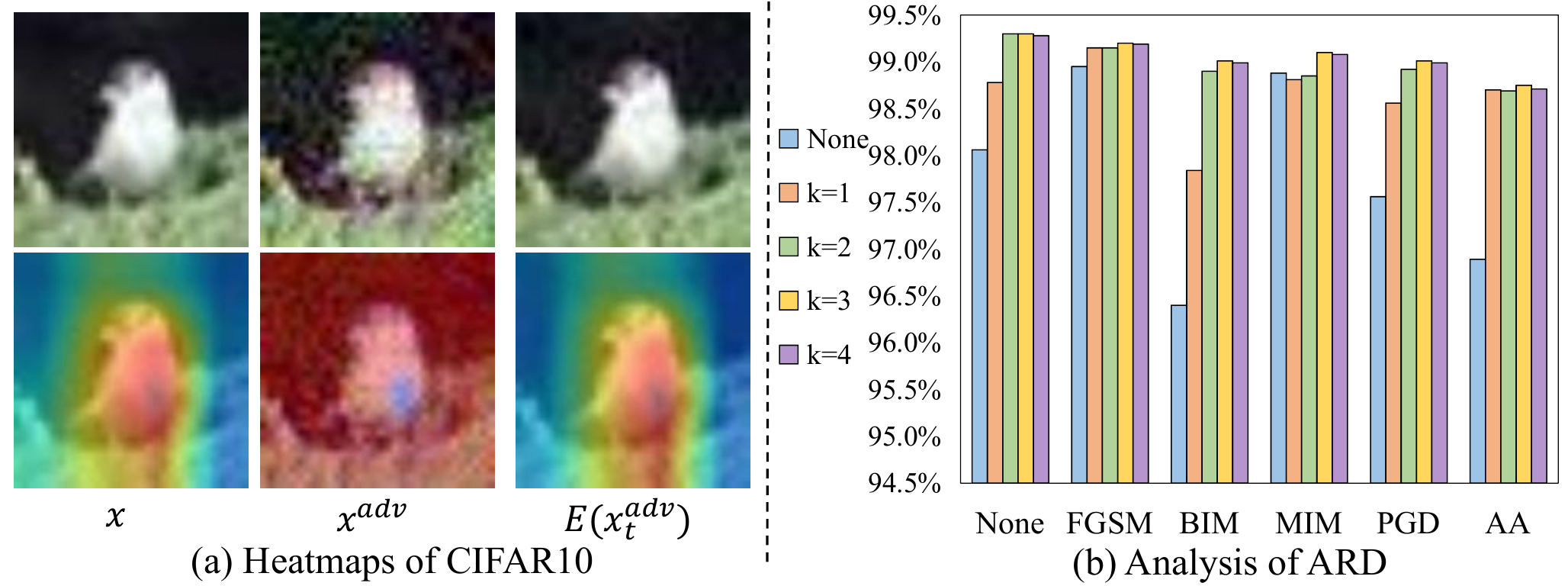} 
\caption{Grad-CAMs and dropout times analysis in Eq.\ref{eq11}.}
\label{fig_K}
\vspace{-4.0mm}
\end{figure}

\textbf{Feature Visualization}. Fig. \ref{fig_tsne} visualizes the feature distribution of the penultimate layer. We use t-SNE \cite{van2008visualizing} to project the features of CIFAR10 onto a two-dimensional plane, where the plots in the $\rm 1^{st}$ row are from the baseline and the $\rm 2^{nd}$ row indicates the proposed PSSR. We observe that the adversarial attack tends to distort the discriminative feature distribution of the baseline model, while PSSR can well restore the discriminative healthy distribution with better feature clusters (i.e., inter-class separability and intra-class compactness) from the adversarial distributions. This verifies that the proposed CAE conforms to the third condition: \textit{consistent with semantic cognition of the target model}. 

\textbf{Grad-CAM Visualization.} Fig.~\ref{fig_K}(a) shows the image $x$, its adversarial sample and the regenerated image together with their heatmaps. We observe the regenerated image is consistent with the raw image in appearance and heatmaps.

\textbf{Analysis of Eq.\ref{eq11} in ARD.} As shown in Eq.\ref{eq11}, we consider randomly dropping out features twice. By setting the dropout times as $k$, we discuss different $k$-value in Fig.~\ref{fig_K}(b). We observe $k=2$ is far better than $k=1$, which shows the effect of AR-Drop. Larger $k$ does not show clear gain but needs much computation cost.

\section{Conclusion}
We have two critical findings that adversarial attack can be decomposed into salient and trivial components, and the salient component dominates attack while the trivial component shows generalizable robustness.
Based on the findings, inspired by the targeted therapies for cancer, we propose a two-stage defense model, Pixel-Surgery and Semantic Regeneration (PSSR). The former aims to remove salient component and retain trivial component, while the latter aims to restore the healthy natural semantics from the residual trivial component with a novel conditional aligned extrapolator (CAE). 
Numerous experiments and analysis demonstrate the proposed model has comparable adversarial robustness but also sufficient interpretability. 

{
    \small
    \bibliographystyle{ieeenat_fullname}
    \bibliography{main}
}


\end{document}